# Component-Enhanced Chinese Character Embeddings

**Yanran Li**[1], **Wenjie Li**[1], **Fei Sun**[2], and **Sujian Li**[3]

[1]Department of Computing, The Hong Kong Polytechnic University, Hong Kong
[2]Institute of Computing Technology, Chinese Academy of Sciences, China
[3]Key Laboratory of Computational Linguistics, Peking University, MOE, China
{csyli, cswjli}@comp.polyu.edu.hk, ofey.sunfei@gmail.com,
lisujian@pku.edu.cn

## Abstract

Distributed word representations are very useful for capturing semantic information and have been successfully applied in a variety of NLP tasks, especially on English. In this work, we innovatively develop two component-enhanced Chinese character embedding models and their bigram extensions. Distinguished from English word embeddings, our models explore the compositions of Chinese characters, which often serve as semantic indictors inherently. The evaluations on both word similarity and text classification demonstrate the effectiveness of our models.

## 1 Introduction

Due to its advantage over traditional one-hot representation, distributed word representation has demonstrated its benefit for semantic representation in various NLP tasks. Among the existing approaches (Huang et al., 2012; Levy and Goldberg, 2014; Yang and Eisenstein, 2015), the continuous bag-of-words model (CBOW) and the continuous skip-gram model (SkipGram) remain the most popular ones that one can use to build word embeddings efficiently (Mikolov et al., 2013a; Mikolov et al., 2013b). These two models learn the distributed representation of a word based on its context. The context defined by the window of surrounding words may unavoidably include certain less semantically-relevant words and/or miss the words with important and relevant meanings (Levy and Goldberg, 2014).

To overcome this shortcoming, a line of research deploys the order information of the words in the contexts by either deriving the contexts using dependency relations where the target word participates (Levy and Goldberg, 2014; Yu and Dredze, 2014; Bansal et al., 2014) or directly keeping the order features (Ling et al., 2015). As to another line, Luong et al. (2013) captures morphological composition by using neural networks and Qiu et al. (2014) introduces the morphological knowledge as both additional input representation and auxiliary supervision to the neural network framework. While most previous work focuses on English, there is a little work on Chinese. Zhang et al. (2013) extracts the syntactical morphemes and Cheng et al. (2014) incorporates the POS tags and dependency relations. Basically, the work in Chinese follows the same ideas as in English.

Distinguished from English, Chinese characters are logograms, of which over 80% are phono-semantic compounds, with a semantic component giving a broad category of meaning and a phonetic component suggesting the sound[1]. For example, the semantic component 亻 (*human*) of the Chinese character 他 (*he*) provides the meaning connected with human. In fact, the components of most Chinese characters inherently bring with certain levels of semantics *regardless of* the contexts. Being aware that the components of Chinese characters are finer grained semantic units, then an important question arises before slipping to the applications of word embeddings—would it be better to learn the semantic representations from the character components in Chinese?

We approach this question from both the practical and the cognitive points of view. In practice, we expect the representations to be optimized for good generalization. As analyzed before, the components are more generic unit *inside* Chinese characters that provides semantics. Such *inherent* information somehow alleviates the shortcoming of the *external* contexts. From the cognitive point of view, it has been found that the knowledge of semantic components significantly corre-late to Chinese word reading and sentence comprehension (Ho et al., 2003).

---
[1]http://en.wikipedia.org/wiki/Radical_(Chinese_characters)

These evidences inspire us to explore novel Chinese character embedding models. Different from word embeddings, character embeddings relate Chinese characters that occur in similar contexts with their component information. Chinese characters convey the meanings from their components, and beyond that, the meanings of most Chinese words also take roots in their composite characters. For example, the meaning of the Chinese word 摇篮 (*cradle*) can be interpreted in terms of its composite characters 摇 (*sway*) and 篮 (*basket*). Considering this, we further extend character embeddings from uni-gram models to bi-gram models.

At the core of our work is the exploration of Chinese semantic representations from a novel character-based perspective. Our proposed Chinese character embeddings incorporate the finer-grained semantics from the components of characters and in turn enrich the representations inherently in addition to utilizing the external contexts. The evaluations on both intrinsic word similarity and extrinsic text classification demonstrate the effectiveness and potentials of the new models.

## 2 Component-Enhanced Character Embeddings

Chinese characters are often composed of smaller and primitive **components** called radicals or radical-like components, which serve as the most basic units for building character meanings. Dating back to the 2nd century AD, the Han dynasty scholar Shen XU organizes his etymological dictionary *shuō wén jiě zì* (*word and expression*) by selecting 540 recurring graphic components that he called bù (means "categories"). Bù is nearly the same as what we call **radicals** today[2]. Most radicals are common semantic components. Over time, some original radicals evolve into radical-like components. Nowadays, a Chinese character often contains exactly one radical (rarely has two) and several other radical-like components. In what follows, we refer to as components both radicals and radical-like components.

Distinguished from English, these composite components are unique and inherent features inside Chinese characters. A lot of times, they allow us to assumingly understand or infer the meanings of characters without any context. In other words, the component-level features inherently bring with additional information that benefits semantic representations of characters. For example, we know that the characters 你 (*you*), 他 (*he*), 伙 (*companion*), 侣 (*companion*), and 们 (*people*) all have the meanings related to human because of their shared component 亻 (*human*), a variant of the Chinese character 人 (*human*). This kind of component information is intrinsically different from the contexts deriving by dependency relations and POS tags. It motivates us to investigate the component-enhanced Chinese character embedding models. While Sun et al. (2014) utilizes radical information in a supervised fashion, we build our models in a holistic unsupervised and bottom-up way.

It is important to note the variation of a radical inside a character. There are two types of variations. The main type is position-related. For example, the radical of the Chinese character 水 (*water*) is itself, but it becomes 氵 as the radical of 池 (*pool*). The original radicals are stretched or squeezed so that they can fit into the general Chinese character shape of a square. The second variation type emerges along with the history of character simplification when traditional characters are converted into simplified characters. For instance, 食 (*eat*) is written as 飠 when it forms as a part of some traditional characters, but is written as 饣 in simplified characters. To cope with these variations and recover the semantics, we match all the radical variants back into their original forms. We extract all the components to build a **component list** for each Chinese character. With the assumption that a character's radical often bring more important semantics than the rest[3], we regard the radical of a character as the first component in its component list.

Let a sequence of characters $D = \{z_1, \ldots, z_N\}$ denotes a corpus of $N$ characters over the character vocabulary $V$. And $z, c, e, K, T, M, |V|$ denote the Chinese character, the context character, the component list, the corresponding embedding dimension, the context window size, the number of components taken into account for each character, and the vocabulary size, respectively. We develop two component-enhanced character embedding models, namely *char*CBOW

---

[2] http://en.wikipedia.org/wiki/Radical_(Chinese_characters)

[3] Inside a character, its radical often serves as the semantic-component while its other radical-like components may be phonetics.

and *char*SkipGram.

*char*CBOW follows the original continuous bag-of-words model (CBOW) proposed by (Mikolov et al., 2013a). We predict the central character $z_i$ conditioned on a $2(M+1)TK$-dimensional vector that is the concatenation of the remaining character-level contexts $(c_{i-T}, \ldots, c_{i-1}, c_{i+1}, \ldots, c_{i+T})$ and the components in their component lists. More formally, we wish to maximize the log likelihood of all the characters as follows,

$$L = \sum_{z_i^n \in D} \log p(z_i|h_i),$$
$$h_i = \text{cat}(c_{i-T}, e_{i-T}, \ldots, c_{i+T}, e_{i+T})$$

where $h_i$ denotes the concatenation of the component-enhanced contexts. We make prediction using a $2KT(M+1)|V|$-dimensional matrix **O**. Different from the original CBOW model, the extra parameter introduced in the matrix **O** allows us to maintain the relative order of the components and treat the radical differently from the rest components.

The development of *char*SkipGram is straightforward. We derive the component-enhanced contexts as $(\langle c_{i-T}, e_{i-T} \rangle, \ldots, \langle c_{i+T}, e_{i+T} \rangle)$ based on the central character $z_i$. The sum of log probabilities given $z_i$ is maximized:

$$L = \sum_{z_i \in D} \sum_{\substack{j=-T \\ j \neq 0}}^{T} \Big( \log p(c_{j+i}|z_i) + \log p(e_{j+i}|z_i) \Big)$$

Figure 1 illustrates the two component-enhanced character embedding models. It is easy to extend *char*CBOW and *char*SkipGram to their corresponding bi-character extensions. Denote the $z_i$, $c_i$ and $e_i$ in *char*CBOW and *char*SkipGram as uni-character $z_{ui}$, $c_{ui}$ and $e_{ui}$, the bi-character extensions are the models fed by bi-character formed $z_{bi}$, $c_{bi}$ and $e_{bi}$.

## 3 Evaluations

We examine the quality of the proposed two Chinese character embedding models as well as their corresponding extensions on both intrinsic word similarity evaluation and extrinsic text classification evaluation.

**Word Similarity**. As the widely used public word similarity datasets like WS-353 (Finkelstein et al., 2001), RG-65 (Rubenstein and Goodenough, 1965) are built for English embeddings, we start from developing appropriate Chinese synonym sets. Two candidate choices are Chinese dictionaries HowNet (Dong and Dong, 2006) and HIT-CIR's Extended Tongyici Cilin (denoted as E-TC)[4]. As HowNet contains less modern words, such as 谷歌 (*Google*), we select E-TC as our benchmark for word similarity evaluation.

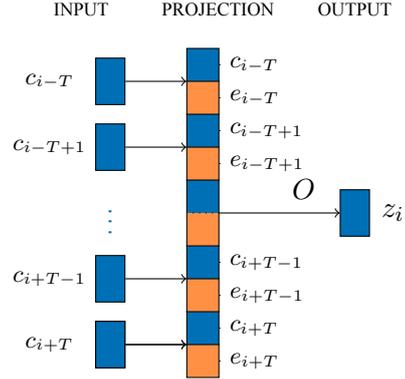

(a) *char*CBOW

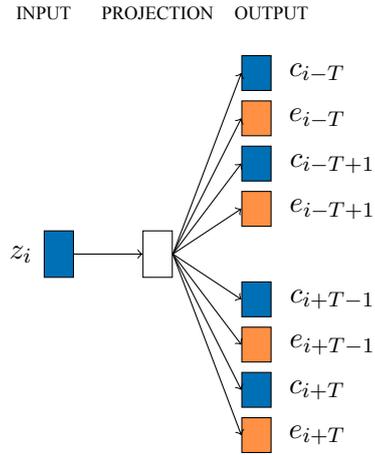

(b) *char*SkipGram

Figure 1: Illustrations of two component-enhanced character embedding models.

**Text Classification**. We use Tencent news titles as our text classification dataset[5]. A total of 8,826 titles of four categories (*society, entertainment, healthcare*, and *military*) are extracted. The lengths of titles range from 10 to 20 words. We train $\ell_2$-regularized logistic regression classifiers using the LIBLINEAR package (Fan et al., 2008) with the learned embeddings.

To build the component-enhanced character embeddings, we employ the GB2312 character set and extract all their component lists. It is easy to obtain the first components (*i.e.*, the radicals), as they are readily available in the *online Xinhua Dictionary*[6]. For the rest radical-like components, we extract them by matching the patterns like "从 (*from*)+X" in the Xinhua dictionary. Such a pattern indicates that a character has a component of X. We also enrich the component lists by matching the pattern "X is only seen" in *Hong Kong Computer Chinese Basic Component Reference*[7].

---

[4] http://ir.hit.edu.cn/demo/ltp/Sharing_Plan.htm
[5] http://www.datatang.com/data/44341

Table 1: Word Similarity Results of Embedding Models

| Model | Spearman's rank correlation (%) | | | | | | | | | | | |
|---|---|---|---|---|---|---|---|---|---|---|---|---|
| | A | B | C | D | E | F | G | H | I | J | K | L |
| CBOW | 33.2 | 25.2 | 32.2 | 27.8 | 36.5 | 37.6 | 43.2 | 40.2 | 37.3 | 39.5 | 44.2 | 40.4 |
| SkipGram | **35.9** | **26.7** | 33.8 | **29.9** | 36.6 | 40.2 | 45.3 | 44.3 | 39.0 | 41.2 | 46.9 | 43.0 |
| *char*CBOW | 34.0 | 23.2 | **34.1** | 26.7 | **37.8** | **49.2** | **48.1** | **44.5** | **40.2** | **42.0** | **48.0** | **43.2** |
| *char*SkipGram | 33.8 | 22.6 | 33.1 | 25.2 | 37.2 | 47.5 | 48.0 | 43.0 | 38.8 | 40.9 | 46.5 | 41.8 |
| CBOW-bi | 37.0 | 27.8 | 34.2 | 29.2 | 38.1 | 43.2 | 50.3 | 48.2 | 43.5 | 46.3 | 50.9 | 45.2 |
| SkipGram-bi | **38.2** | **29.0** | 34.0 | 29.4 | 38.9 | 44.9 | 50.2 | 49.3 | **45.6** | 48.4 | 51.3 | 47.4 |
| *char*CBOW-bi | 36.0 | 25.3 | **36.8** | **31.2** | **40.2** | **54.3** | **55.7** | **49.7** | 45.3 | **48.9** | **53.2** | **47.7** |
| *char*SkipGram-bi | 35.7 | 24.6 | 33.4 | 30.5 | 39.7 | 53.3 | 53.9 | 48.2 | 33.2 | 47.1 | 52.0 | 45.7 |

Table 2: Text Classification Results of Embedding Models

| Model | Society | | | Entertainment | | | Healthcare | | | Military | | |
|---|---|---|---|---|---|---|---|---|---|---|---|---|
| | P | R | F | P | R | F | P | R | F | P | R | F |
| CBOW-bi | 43.0 | 28.0 | 33.9 | 48.2 | 32.7 | 39.0 | 47.6 | 29.5 | 36.4 | 57.6 | 40.8 | 47.8 |
| SkipGram-bi | 47.2 | 31.1 | 37.5 | 49.8 | 34.0 | 40.4 | 48.4 | 32.7 | 39.0 | 58.8 | 42.3 | 49.2 |
| *char*CBOW-bi | **57.4** | **37.4** | **45.2** | **62.2** | **42.0** | **50.1** | **59.2** | **45.3** | **51.3** | **70.3** | **51.0** | **59.1** |
| *char*SkipGram-bi | 50.3 | 34.6 | 41.0 | 57.6 | 34.5 | 43.2 | 57.3 | 42.5 | 48.8 | 67.8 | 48.3 | 56.4 |
| CBOW-combine | 46.2 | 29.0 | 35.6 | 50.3 | 35.0 | 41.3 | 51.0 | 33.6 | 40.5 | 62.2 | 45.7 | 52.7 |
| SkipGram-combine | 50.9 | 34.6 | 41.2 | 51.4 | 37.2 | 43.2 | 52.1 | 35.6 | 42.3 | 62.1 | 49.0 | 54.8 |
| *char*CBOW-combine | **62.2** | **39.8** | **48.5** | **66.7** | **46.6** | **54.9** | **62.2** | **50.2** | **55.6** | **74.4** | **53.8** | **62.4** |
| *char*SkipGram-combine | 54.4 | 38.2 | 44.9 | 59.2 | 36.5 | 45.2 | 62.0 | 47.9 | 54.0 | 73.4 | 53.5 | 61.9 |

It is observed that nearly 65% Chinese characters have only one component (their radicals), and 95% Chinese characters have two components (including their radicals). Thus, we decide to maintain up to two extracted components to build the character embeddings according to the frequency of their occurrences. To cope with the radical variation problem, we transform 24 radical variants to their origins, such as 亻 to 人 (*human*), 扌 to 手 (*hand*), 氵 to 水 (*water*) and 辶 to 辵 (*foot*). The complete list of the transformations is provided in Appendix for easy reference.

We adopt Chinese Wikipedia Dump[8] to train our models as well as the original CBOW and SkipGram, implemented in the `Word2Vec` tool[9] for comparison. The corpus in total contains 232,894 articles. In preprocessing, we remove pure digit words and non-Chinese characters, and ignore the words less than 10 occurrences during training. We set the context window size $T$ as 2 and use 5 negative samples experimentally. All the embedding dimensions $K$ are set to 50.

In the word similarity evaluation, we compute the Spearman's rank correlation (Myers et al., 2010) between the similarity scores based on the learned embedding models and the E-TC similarity scores computed by following Jiu-le and Wei (2010). The bi-character embeddings are concatenation of the composite character embeddings. For the text classification evaluation, we average the composite single character embeddings for each bi-gram. And each bi-gram overlaps with the previous one. The titles are represented by averaging

---
[6] http://xh.5156edu.com/
[7] http://www.ogcio.gov.hk/tc/business/tech_promotion/ccli/cliac/glyphs_guidelines.htm
[8] http://download.wikipedia.com/zhwiki/
[9] https://code.google.com/p/word2vec/

the embeddings of their composite grams[10].

Table 1 presents the word similarity evaluation results of the eight embedding models mentioned above, where A–L denote the twelve categories in E-TC. The first four rows are the results with the uni-character inputs, and the last four rows correspond to the bi-character embeddings results.

We can see that both CBOW and CBOW-bi perform worse than the corresponding SkipGram and SkipGram-bi. This result is consistent with the finding in the previous work (Pennington et al., 2014; Levy and Goldberg, 2014; Levy et al., 2015). To some extent, CBOW and its extension CBOW-bi are the most different among the eight (the first four models in Table 1 and the first four models in Table 2). They tie together the characters in each context window by representing the context vector as the sum of their characters' vectors. Although they have a potential of deriving better representations (Levy et al., 2015), they lose some particular information from each unit of input in the average operations.

Although the performance on twelve different categories varies, in overall *char*CBOW, *char*SkipGram and their extensions consistently better correlate to E-TC. It provides the evidence that the component information in Chinese characters is of significance. Clearly, the bi-character models achieve higher rank correlations. These results are not surprised. As a matter of fact, a majority of Chinese words are compounds of two characters. Thus, in many cases two characters together is equivalent to a Chinese word. Considering the superiority of the bi-character models, we only apply them in the text classification evaluations.

The results shown in the first four rows of Table 2 are similar to those in the word similarity evaluation. Please notice the significant improvement of *char*CBOW and *char*CBOW-bi. We conjecture this as a hint of the importance of the order information, which is introduced by the extra parameter in the output matrixes. Their better performances verify our assumption that the radicals are more important than non-radicals. This is also attributed to the benefit from the order of the characters in the contexts.

Actually, we also conduct an additional experiment to combine the uni-gram and the bi-gram embeddings for text classification and notice in average about 8.4% of gain over the bi-gram embeddings alone. The detailed results are presented in the last four rows of Table 2.

## 4 Conclusions and Future Work

In this paper, we propose two component-enhanced Chinese character embedding models and their extensions to explore both the internal compositions and the external contexts of Chinese characters. Experimental results demonstrate their benefits in learning rich semantic representations. For the future work, we plan to devise embedding models based together on the composition of component-character and of character-word. The two types of compositions will serve in a coordinate fashion for the distributional representations.

## Acknowledgements

The work described in this paper was supported by the grants from the Research Grants Council of Hong Kong (PolyU 5202/12E and PolyU 152094/14E), the grants from the National Natural Science Foundation of China (61272291 and 61273278) and a PolyU internal grant (4-BCB5).

## Appendix

As mentioned in Section 3, we present the complete list of transformations of the variant and original forms of 24 radicals. The *meaning* columns provide the corresponding meanings of the components in the left.

| transform | meaning | transform | meaning |
|---|---|---|---|
| 艹 → 艸 | grass | 扌 → 手 | hand |
| 亻 → 人 | human | 氵 → 水 | water |
| 刂 → 刀 | knife | 車 → 车 | vehicle |
| 犭 → 犬 | dog | 攵 → 攴 | hit |
| 灬 → 火 | fire | 纟 → 糹 | silk |
| 钅 → 金 | gold | 耂 → 老 | old |
| 麥 → 麦 | wheat | 牜 → 牛 | cattle |
| 饣 → 食 | eat | 飠 → 食 | eat |
| 礻 → 示 | memory | 忄 → 心 | heart |
| 罒 → 网 | nest | 王 → 玉 | jade |
| 讠 → 言 | speak | 衤 → 衣 | cloth |
| 月 → 肉 | body | 辶 → 辵 | walk |

---

[10] We do not compare the uni-formed characters with bi-formed compound characters. The word pairs that cannot be found in the vocabulary are removed.

# References

Mohit Bansal, Kevin Gimpel, and Karen Livescu. 2014. Tailoring continuous word representations for dependency parsing. In *Proceedings of the Annual Meeting of the Association for Computational Linguistics*.

Fei Cheng, Kevin Duh, and Yuji Matsumoto. 2014. Parsing chinese synthetic words with a character-based dependency model.

Zhendong Dong and Qiang Dong. 2006. *HowNet and the Computation of Meaning*. World Scientific.

Rong-En Fan, Kai-Wei Chang, Cho-Jui Hsieh, Xiang-Rui Wang, and Chih-Jen Lin. 2008. Liblinear: A library for large linear classification. *The Journal of Machine Learning Research*, 9:1871–1874.

Lev Finkelstein, Evgeniy Gabrilovich, Yossi Matias, Ehud Rivlin, Zach Solan, Gadi Wolfman, and Eytan Ruppin. 2001. Placing search in context: The concept revisited. In *Proceedings of the 10th international conference on World Wide Web*, pages 406–414. ACM.

Connie Suk-Han Ho, Ting-Ting Ng, and Wing-Kin Ng. 2003. A ?radical? approach to reading development in chinese: The role of semantic radicals and phonetic radicals. *Journal of Literacy Research*, 35(3):849–878.

Eric H Huang, Richard Socher, Christopher D Manning, and Andrew Y Ng. 2012. Improving word representations via global context and multiple word prototypes. In *Proceedings of the 50th Annual Meeting of the Association for Computational Linguistics: Long Papers-Volume 1*, pages 873–882. Association for Computational Linguistics.

TIAN Jiu-le and ZHAO Wei. 2010. Words similarity algorithm based on tongyici cilin in semantic web adaptive learning system [j]. *Journal of Jilin University (Information Science Edition)*, 6:010.

Omer Levy and Yoav Goldberg. 2014. Dependencybased word embeddings. In *Proceedings of the 52nd Annual Meeting of the Association for Computational Linguistics*, volume 2, pages 302–308.

Omer Levy, Yoav Goldberg, and Ido Dagan. 2015. Improving distributional similarity with lessons learned from word embeddings. *Transactions of the Association for Computational Linguistics*, 3:211–225.

Wang Ling, Chris Dyer, Alan Black, and Isabel Trancoso. 2015. Two/ too simple adaptations of word2vec for syntax problems. In *Proceedings of the 2015 Conference of the North American Chapter of the Association for Computational Linguistics: Human Language Technologies*. Association for Computational Linguistics.

Minh-Thang Luong, Richard Socher, and Christopher D Manning. 2013. Better word representations with recursive neural networks for morphology. *CoNLL-2013*, 104.

Tomas Mikolov, Kai Chen, Greg Corrado, and Jeffrey Dean. 2013a. Efficient estimation of word representations in vector space. *arXiv preprint arXiv:1301.3781*.

Tomas Mikolov, Ilya Sutskever, Kai Chen, Greg S Corrado, and Jeff Dean. 2013b. Distributed representations of words and phrases and their compositionality. In *Advances in neural information processing systems*, pages 3111–3119.

Jerome L Myers, Arnold Well, and Robert Frederick Lorch. 2010. *Research design and statistical analysis*. Routledge.

Jeffrey Pennington, Richard Socher, and Christopher D Manning. 2014. Glove: Global vectors for word representation. *Proceedings of the Empiricial Methods in Natural Language Processing (EMNLP 2014)*, 12:1532–1543.

Siyu Qiu, Qing Cui, Jiang Bian, Bin Gao, and Tie-Yan Liu. 2014. Co-learning of word representations and morpheme representations. COLING.

Herbert Rubenstein and John B Goodenough. 1965. Contextual correlates of synonymy. *Communications of the ACM*, 8(10):627–633.

Yaming Sun, Lei Lin, Nan Yang, Zhenzhou Ji, and Xiaolong Wang. 2014. Radical-enhanced chinese character embedding. In *Neural Information Processing*, pages 279–286. Springer.

Yi Yang and Jacob Eisenstein. 2015. Unsupervised multi-domain adaptation with feature embeddings.

Mo Yu and Mark Dredze. 2014. Improving lexical embeddings with semantic knowledge. In *Association for Computational Linguistics (ACL)*, pages 545–550.

Meishan Zhang, Yue Zhang, Wanxiang Che, and Ting Liu. 2013. Chinese parsing exploiting characters. In *ACL (1)*, pages 125–134.